\documentclass[11pt]{article}

\usepackage[margin=1in]{geometry}
\usepackage{amsmath,amssymb}
\usepackage{booktabs}
\usepackage{hyperref}
\usepackage{xcolor}
\usepackage{enumitem}
\usepackage{microtype}
\usepackage{adjustbox}
\usepackage{pgfplots}
\pgfplotsset{compat=1.18}
\usetikzlibrary{arrows.meta,positioning}
\usepackage{placeins}

\hypersetup{
  colorlinks=true,
  linkcolor=blue,
  citecolor=blue,
  urlcolor=blue
}

\title{\textbf{Spike-Aware INT8 Execution for Spiking Language Models on Commodity CPUs}}

\author{
  Ting Liu\\
  SymbolicLight Research\\
  Foshan, Guangdong, China\\
  \texttt{research@symboliclight.com}
}

\date{June 2026 (revised September 2026)}

\begin{document}
\maketitle

\begin{abstract}
Binary spike activations allow a language-model runtime to read only active weight columns and replace multiplications by weight sums. We implement this execution strategy in C++ for an 874M-parameter spike-gated language model. Sparse projections use column-major INT8 weights, integer accumulation, and one scale application per output channel; dense projections retain row-major access and FP32 activations. In a single-thread comparison using an early checkpoint, INT8 achieves 23.31 tokens/s versus 9.82 for FP32, while reducing weight storage from 3355.2 to 1087.4 MiB. A variant using INT4 on dense projections saves a further 17.4\% of storage but reduces decode throughput by 46.6\%. On an AMD Ryzen 7 5800X, the final INT8 checkpoint achieves 22.63 tokens/s on one thread and 47.90 on four threads; 512-token prefill reaches 94.68 tokens/s on eight threads. A separate ARM output-head case study records higher trimmed decode-window energy metrics for two candidate-verification configurations. The results characterize how activation-specific layouts and quantized kernels support CPU deployment of a spike-gated language model.
\end{abstract}

\section{Introduction}

A binary spike vector changes the computation required by a language-model projection. For a weight matrix $W$ and an input $x\in\{0,1\}^K$, the product $Wx$ is the sum of columns whose input spike is active. A CPU runtime can therefore skip inactive columns and avoid multiplying active weights by one. Realizing this execution pattern requires the weight layout and accumulation kernel to follow the activation structure.

The SymbolicLight V1 language model combines Leaky Integrate-and-Fire (LIF) spike dynamics with a continuous residual stream~\cite{symboliclightv1}. Its binary and continuous paths require different projection kernels: sparse column reads serve the former, while contiguous row-wise dot products serve the latter. We implement this split in C++, with per-output-channel INT8 weights and integer accumulation on binary inputs. Weight loading and tokenization are independent of PyTorch at inference time.

The paper develops two contributions. The first is an execution design that connects binary spike states, active-column storage, and integer weight sums while preserving dense-path locality. The second is a deployment evaluation of weight precision, memory use, and CPU thread scaling. On an early checkpoint, the tested INT8 implementation decodes at 2.37 times the FP32 rate with 32.4\% of its weight storage. On the final 874M-parameter checkpoint, one- and four-thread decode reach 22.63 and 47.90 tokens/s on Ryzen 7 5800X. These measurements provide concrete operating points for a language module sharing a CPU with other local workloads.

\paragraph{Relation to existing inference systems.}
Dense Transformer models~\cite{vaswani2017attention} motivate runtimes such as llama.cpp, which combines quantization, the GGUF model format, and hardware-specific kernels~\cite{llamacpp}. BitNet and bitnet.cpp explore low-bit weights and ternary inference~\cite{bitnet,bitnetcpp}. Our design uses the model's binary \emph{activations} to select which weight columns are read; weight precision and activation sparsity jointly determine its execution path.

\paragraph{Relation to spiking language models.}
Spiking neuron models and surrogate-gradient training provide the background for learning with discrete events~\cite{gerstner2002spiking,neftci2019surrogate}. SpikeGPT applies spiking mechanisms to language modeling~\cite{zhu2023spikegpt}; SymbolicLight V1 reports a dual-path architecture with 194M-parameter pre-training experiments and a sub-billion-parameter scale-up~\cite{symboliclightv1}. The present study addresses execution of that model family on conventional CPUs. A separate ARM case study examines candidate-row verification in the output head, alongside the main CPU evaluation.

\section{Model and Execution Design}
\label{sec:design}

\subsection{Binary and Continuous Paths}

The evaluated decoder has 22 blocks, a 1536-dimensional residual stream, and 874M parameters. Each block combines binary spike states with continuous hidden states. The sequence mixer, named ternary content-addressable memory (TCAM) in the model, includes a spike-conditioned projection and a recurrent decay path, together with an attention path operating on the continuous stream. Their outputs are gated and fused. Thus, the implementation retains query, key, and value projections and a key/value cache.

The two large binary-input projections are the TCAM projection and the feed-forward down projection. Query/key/value and attention-output projections, the feed-forward up projection, and the vocabulary output projection operate on continuous inputs. The runtime assigns layouts and kernels at this operator boundary (Figure~\ref{fig:execution}).

Two checkpoints of this architecture are used. The \emph{early checkpoint}, after 30,000 training steps, supplies the precision comparison and historical layout measurements. The \emph{final checkpoint}, after 186,000 steps, supplies the final CPU evaluation, final-checkpoint numerical checks, and ARM case study. Early-checkpoint numerical checks are reported separately. The deployed configuration has 57,344 output vocabulary slots and a SentencePiece tokenizer with 55,296 entries; this is the tokenizer used by the runtime measurements, distinct from the 48K vocabulary described in the architecture report.

\subsection{Active-Column Execution}

For a binary input, let $A=\{k:x_k=1\}$ be the active index set. The runtime gathers these indices and stores the corresponding projection matrix column-major. Each selected column is then a contiguous read across output channels. Inactive columns are not read by the projection kernel. Dense projections remain row-major, so each output is computed from a contiguous dot product and written once.

This division avoids imposing the sparse accumulator's access pattern on dense operators. In the historical FP32 layout experiments, all-column-major execution achieved 8.2 tokens/s, versus 10.4 for all-row-major AVX2 and 13.0 for the mixed layout. Appendix~\ref{app:history} preserves the development measurements and their timing conditions.

\begin{figure}[!htbp]
\centering
\begin{tikzpicture}[box/.style={draw=black!60,rounded corners=2pt,align=center,font=\small,minimum height=1cm,text width=3.7cm},flow/.style={-{Stealth[length=2mm]},thick}]
\node[box,fill=blue!5] (s) {Binary spikes\\$x_k\in\{0,1\}$\\Active indices $A$};
\node[box,right=0.5cm of s,fill=blue!5] (c) {Read active columns};
\node[box,right=0.5cm of c,fill=blue!5] (i) {Integer sum\\then row scale\\$y_m=s_m\sum_{k\in A}Q_{m,k}$};
\node[box,below=0.65cm of s,fill=gray!8] (d) {FP32 activations};
\node[box,right=0.5cm of d,fill=gray!8] (r) {Read each weight row};
\node[box,right=0.5cm of r,fill=gray!8] (f) {FP32 dot product\\then row scale};
\draw[flow] (s)--(c); \draw[flow] (c)--(i);
\draw[flow] (d)--(r); \draw[flow] (r)--(f);
\end{tikzpicture}
\caption{Projection execution by activation type. The upper path reads only active input columns and sums integer weights for unit-valued binary activations. The lower path evaluates row-major dot products with continuous activations. Both apply per-output-channel scales after accumulation. Here $A$ is the active input index set, $Q_{m,k}$ is an INT8 weight, $s_m$ is the scale for output channel $m$, and $y_m$ is its output.}
\label{fig:execution}
\end{figure}

\subsection{Quantization and Accumulation}

Most two-dimensional weights use symmetric INT8 quantization with one scale per output channel. For $W\in\mathbb{R}^{M\times K}$,
\[
 s_m=\frac{\max(\max_k|W_{m,k}|,10^{-8})}{127},\qquad
 Q_{m,k}=\mathrm{clip}\bigl(\mathrm{round}(W_{m,k}/s_m),-127,127\bigr).
\]
The positive floor avoids division by zero for an all-zero row. On the binary path, the quantized projection is
\begin{equation}
 y_m=s_m\sum_{k\in A}Q_{m,k}.
\label{eq:sparse}
\end{equation}
The inner loop sums integer weights across active columns. The result is converted to FP32 and scaled once per output channel. The general accumulator uses INT32; bounded active-count cases can use exact lower-width partial sums. Equation~\ref{eq:sparse} relies on unit-valued binary inputs. Continuous inputs use a weight-only INT8 kernel that converts weights to FP32, accumulates products with FP32 activations, and then applies the row scale.

AVX2/FMA kernels vectorize dense row reads, FP32 sparse-column accumulation, and layer normalization. The sparse path vectorizes over output channels while iterating over active indices. Embeddings and normalization parameters retain FP32 storage. The mixed-precision variant, denoted INT4mix, stores dense projections in INT4 and retains INT8 for the two sparse projection classes.

\paragraph{Export and loading.}
The exporter writes tensor shapes, layouts, weight files, and quantization scales to a manifest. The C++ loader parses exact tensor keys and explicit shape metadata, including one-element representations of scalar parameters. This supplies the operator-specific storage used by the kernels. The runtime supports autoregressive generation and native perplexity evaluation with a separately loaded tokenizer.

\section{Experimental Setup and Numerical Validation}
\label{sec:setup}

\subsection{CPU Workloads and Measurements}

CPU benchmarks use an AMD Ryzen 7 5800X (8 cores, 16 hardware threads) under Windows with AVX2/FMA support. The precision comparison uses the early checkpoint and one thread. Final-checkpoint measurements use the per-channel INT8 export. Decode runs generate up to 100 tokens from ``The quick brown fox jumps over the lazy dog.'' with temperature 0 and seed 42. OpenMP controls the runtime's thread count. Each setting is launched three times; results report the median and minimum--maximum range. Generated token IDs agree across the repeated trials at each thread count.

Weight sizes and resident set size (RSS) are reported in MiB ($2^{20}$ bytes). Peak RSS is the median of the three per-run peaks, sampled approximately every 50 ms over the process and its children; peaks shorter than the sampling interval can be missed. Prefill timing uses a beginning-of-sequence token followed by repeated copies of token ID 100 to obtain lengths of 16, 64, 128, 256, and 512 tokens. Each length/thread setting has three trials and one generated token.

Dense CPU timing references and their evaluation protocols are retained in Appendix~\ref{app:dense}. They use llama.cpp's synthetic benchmark workload rather than the text-generation workload above. The main evaluation therefore compares precision variants and thread settings within the spiking runtime. The x86 measurements cover throughput and memory on this host; ARM power measurements are described with their case study in Section~\ref{sec:power}.

\subsection{Numerical Validation}

Table~\ref{tab:alignment} separates implementation agreement from the change introduced by quantization. The final-checkpoint FP32 configuration agrees with its PyTorch reference on the three tested inputs. For INT8, a reference built from the same dequantized export removes the difference in weight values: the 2- and 8-token checks pass, while the 60-token check has mean absolute error 0.688945 and maximum error 1.99516 despite matching top-1. The source of this discrepancy is unresolved. The comparison against FP32 weights includes both quantization differences and implementation differences.

These July checks describe the listed configurations and inputs; they do not establish numerical equivalence for the earlier benchmark executable or for arbitrary INT8 contexts. The performance results below are measurements of the tested runtime configurations. An early-checkpoint check recorded mean/max absolute logit errors of 0.00528762/0.0101566 and matching top-5 identities. Multilingual generation smoke tests additionally checked loading, tokenization, and output generation.

\begin{table}[!htbp]
\centering
\small
\caption{C++/PyTorch logit checks for the final checkpoint, recorded in July 2026. Context length is in tokens; errors are over the output vocabulary. Pass means maximum absolute error at most the recorded threshold of 0.01; top-1 denotes agreement on the highest-scoring token. The first and third groups use FP32 key/value caches and exact, error-function GELU in C++.}
\label{tab:alignment}
\begin{adjustbox}{max width=\linewidth}
\begin{tabular}{rrrrr}
\toprule
Context & Mean absolute error & Maximum absolute error & Pass & Top-1 match \\
\midrule
\multicolumn{5}{l}{FP32 implementation vs. FP32 PyTorch reference} \\
2 & 0.000179427 & 0.000257492 & yes & yes \\
8 & 0.000167772 & 0.000239372 & yes & yes \\
60 & 0.000169822 & 0.000268936 & yes & yes \\
\midrule
\multicolumn{5}{l}{INT8 implementation vs. PyTorch with the same dequantized INT8 export} \\
2 & 0.000181893 & 0.000261307 & yes & yes \\
8 & 0.000213429 & 0.000318527 & yes & yes \\
60 & 0.688945 & 1.99516 & no & yes \\
\midrule
\multicolumn{5}{l}{INT8 implementation vs. FP32 PyTorch reference} \\
2 & 0.746827 & 4.77214 & no & yes \\
8 & 0.615934 & 2.54264 & no & no \\
60 & 2.6345 & 5.35261 & no & yes \\
\bottomrule
\end{tabular}
\end{adjustbox}
\end{table}

\FloatBarrier
\section{CPU Deployment Results}
\label{sec:results}

\subsection{Weight Precision, Storage, and Throughput}
\label{sec:evolution}

Table~\ref{tab:current_ablation} compares three weight formats on the early checkpoint using the same AVX2 executable, one thread, and the decode prompt from Section~\ref{sec:setup}. INT8 reaches 23.31 tokens/s, 2.37 times the FP32 mixed-layout rate of 9.82, while reducing weight storage from 3355.2 to 1087.4 MiB. Sampled peak RSS falls from 3421.4 to 1175.9 MiB. All three INT8 decode trials lie between 23.27 and 23.44 tokens/s.

\begin{table}[!htbp]
\centering
\caption{Weight-format comparison on the early checkpoint: one AVX2 executable, one thread, and the same 100-token generation workload. Entries are medians of three runs; indented rows give minimum--maximum ranges. INT4mix uses INT4 dense projections and INT8 sparse projections. RSS is sampled peak resident set size.}
\label{tab:current_ablation}
\begin{adjustbox}{max width=\linewidth}
\begin{tabular}{lrrrr}
\toprule
Runtime variant & Weight MiB & Prefill tokens/s & Decode tokens/s & RSS MiB \\
\midrule
FP32 mixed layout & 3355.2 & 20.52 & 9.82 & 3421.4 \\
\quad range & & 16.37--21.42 & 9.70--11.94 & \\
INT8 mixed layout & 1087.4 & 34.04 & 23.31 & 1175.9 \\
\quad range & & 25.91--34.16 & 23.27--23.44 & \\
INT4mix & 897.9 & 14.62 & 12.45 & 986.5 \\
\quad range & & 14.45--16.70 & 11.07--12.68 & \\
\bottomrule
\end{tabular}
\end{adjustbox}
\end{table}

INT4mix reduces weight storage by another 17.4\% relative to INT8, from 1087.4 to 897.9 MiB, but lowers decode throughput by 46.6\%, from 23.31 to 12.45 tokens/s. The tested INT8 implementation is therefore the higher-throughput choice, while INT4mix provides a smaller export. This comparison measures the combined formats and kernels; it does not isolate the cost of any single quantization operation.

The three variants share the mixed layout. Their comparison evaluates weight precision and kernel implementation, rather than isolating the layout change. Historical layout measurements appear in Appendix~\ref{app:history}. The previous pure-INT4 result is withdrawn because the sparse export had an incompatible scale layout; Appendix~\ref{sec:int4} documents the defect.

\subsection{Final-Checkpoint Decode and Memory}

The final INT8 checkpoint decodes at 22.63 tokens/s on one thread, with a three-trial range of 22.42--22.83. Its weights occupy 1087.4 MiB and sampled peak RSS is 1175.9 MiB. Figure~\ref{fig:threads} shows the thread-scaling curve: throughput rises to 39.08 tokens/s on two threads and 47.90 on four, then reaches 47.28 on eight and 41.80 on sixteen. Four threads provide the highest median, 2.12 times the single-thread rate; the four- and eight-thread trial ranges overlap. Peak RSS stays between 1175.9 and 1177.0 MiB across thread counts.

This workload obtains most of its measured decode throughput with four threads. Increasing the thread count beyond that point provides no median throughput benefit on the tested processor. Table~\ref{tab:thread_scaling} gives the complete values and trial ranges.

\begin{figure}[!htbp]
\centering
\begin{tikzpicture}
\begin{axis}[width=0.84\linewidth,height=0.46\linewidth,xmode=log,log basis x=2,
xtick={1,2,4,8,16},xticklabels={1,2,4,8,16},xlabel={CPU threads},ylabel={Decode tokens/s},
ymin=0,ymax=56,grid=major,grid style={gray!25},legend pos=north west,legend style={font=\small,cells={anchor=west}},mark size=2pt,thick]
\addplot+[error bars/.cd,y dir=both,y explicit] table[x=x,y=y,y error minus=minus,y error plus=plus] {decode_threads.dat};
\end{axis}
\end{tikzpicture}
\caption{Decode thread scaling of the final INT8 checkpoint on Ryzen 7 5800X. Each setting generates 100 tokens from the same prompt. Points are medians of three trials; bars span the minimum and maximum, not confidence intervals. Complete data appear in Table~\ref{tab:thread_scaling}.}
\label{fig:threads}
\end{figure}

\subsection{Prefill Scaling}

Prefill and decode favor different thread counts in these measurements. For a 512-token synthetic prompt, prefill throughput rises from 29.86 tokens/s on one thread to 94.68 on eight, a 3.17-fold increase (Figure~\ref{fig:prefill}). Across prompt lengths, the eight-thread rate is 3.2--4.1 times the single-thread rate. The 512-token prompt takes approximately 17.1 s to prefill at the single-thread median and 5.4 s at the eight-thread median, making prompt-processing time a material part of this deployment profile.

Single-thread trial ranges are within 1.7 tokens/s. Some parallel settings vary substantially: the 512-token four-thread trials span 62.80--86.81 tokens/s, and the 128-token eight-thread trials span 92.43--126.20. The figure displays these observed ranges; Table~\ref{tab:long_prefill} preserves the full matrix.

\begin{figure}[!htbp]
\centering
\begin{tikzpicture}
\begin{axis}[width=0.84\linewidth,height=0.46\linewidth,xmode=log,log basis x=2,
xtick={1,2,4,8},xticklabels={1,2,4,8},xlabel={CPU threads},ylabel={Prefill tokens/s},
ymin=0,ymax=145,grid=major,grid style={gray!25},legend pos=north west,legend style={font=\small,cells={anchor=west}},mark size=2pt,thick]
\addplot+[error bars/.cd,y dir=both,y explicit] table[x=x,y=y,y error minus=minus,y error plus=plus] {prefill_16.dat};
\addlegendentry{16 tokens}
\addplot+[error bars/.cd,y dir=both,y explicit] table[x=x,y=y,y error minus=minus,y error plus=plus] {prefill_64.dat};
\addlegendentry{64 tokens}
\addplot+[error bars/.cd,y dir=both,y explicit] table[x=x,y=y,y error minus=minus,y error plus=plus] {prefill_128.dat};
\addlegendentry{128 tokens}
\addplot+[error bars/.cd,y dir=both,y explicit] table[x=x,y=y,y error minus=minus,y error plus=plus] {prefill_256.dat};
\addlegendentry{256 tokens}
\addplot+[error bars/.cd,y dir=both,y explicit] table[x=x,y=y,y error minus=minus,y error plus=plus] {prefill_512.dat};
\addlegendentry{512 tokens}
\end{axis}
\end{tikzpicture}
\caption{Prefill thread scaling of the final INT8 checkpoint on Ryzen 7 5800X. Legend entries give synthetic prompt lengths. Points are medians of three trials; bars span the minimum and maximum, not confidence intervals. Complete data appear in Table~\ref{tab:long_prefill}.}
\label{fig:prefill}
\end{figure}

\subsection{Language-Model Evaluation Record}

The final INT8 runtime records WikiText-2 perplexity of 24.7979 (24.80 rounded), using 314,599 tokenized inputs, 313,984 scored tokens, sequence length 512, and stride 512. This is an evaluation of the deployed runtime's output distribution. Token-level perplexity depends on tokenizer and scoring protocol~\cite{pplguide}; the separate dense-model records in Appendix~\ref{app:dense} are not a common quality scale.

\FloatBarrier
\section{Output-Head Power Case Study}
\label{sec:power}

Candidate-row verification offers another way to select weight rows during inference. After an approximate projection over the full vocabulary, the runtime can recompute selected high-scoring rows with FP32 activations. We examine two related configurations on a Radxa Rock 5T board with a Rockchip RK3588 ARM processor, the final INT8 export, and one OpenMP thread. The workload is a fixed Chinese prompt asking for joules per token from a given power and speed, followed by 80 greedily generated tokens. The trace and wall-power configurations differ in both gate threshold and scope.

\subsection{Projection-Only Verification Trace}

The trace configuration applies INT8 activations to the vocabulary output projection and triggers candidate verification when the top-two logit gap is below 0.5. For $K=16,32,64$, verification runs in 54 of 80 trace events. Within those events, it evaluates $54K$ rows. With $V=57{,}344$ vocabulary slots, the row-count reduction relative to verifying all rows is $100(1-K/V)$, or 99.972\%, 99.944\%, and 99.888\%, respectively. This denominator covers the verification projection, which follows the initial full projection.

The remaining top-$K$ verification takes 0.871--1.103\% of head time within the selected events; the initial output projection takes 58.08--62.08\% (Table~\ref{tab:zerotrace}). These are timings after candidate selection, so the verification share does not measure the cost of full-vocabulary verification before skipping or the amount of end-to-end time saved. The trace characterizes the work performed by this configuration.

\subsection{Full-Head Configuration and Wall Power}
\label{sec:setup_power}

The wall-power experiment uses the ARM NEON build and a different candidate: INT8 activations extend to the complete head, including its learned bottleneck branch that adds a vocabulary-level correction to the output logits. A top-two gap below 4.0 triggers FP32-activation recomputation of the $K$ selected final-logit rows. The baseline uses FP32 activations for head computation with the same INT8 weight export. Each candidate ($K=32$ or $64$) has two paired trials, ordered baseline--candidate and then candidate--baseline. All candidate outputs match the paired baseline's 80 token IDs.

Whole-board power is sampled by a Xiaomi smart plug over the local network. The requested interval is 0.5 s and the observed median is approximately 0.68 s; individual readings are integer watts. The analysis trims 2 s from each end of a command window and a decode window. The former includes loading and prefill; the latter is inferred backwards from command completion using the runtime's decode duration. For either retained interval $[a,b]$, the recorded metric is
\[
 J_{\mathrm{win}}=\bar P(b-a)/80,
\]
where $\bar P$ is the arithmetic mean of the in-window readings. Because the denominator includes all 80 tokens after time trimming, this is a trimmed-window energy metric, not complete decode energy per generated token. Percentage changes compare the means of the two run values for each variant.

Table~\ref{tab:wall_power} reports decode-rate changes of $-2.891\%$ and $-3.601\%$, with trimmed decode-window metric changes of $+3.571\%$ and $+4.905\%$. Seven of eight runs have mean decode-window readings of 8.00 W and one has 7.95 W. At nearly unchanged recorded power, the metric increases with the longer retained decode duration. Full-window changes are $-1.490\%$ and $+2.650\%$; per-run values are retained in Table~\ref{tab:wall_abs}.

\begin{table}[!htbp]
\centering
\caption{ARM wall-power comparison for the full-head INT8-activation candidates with gate threshold 4.0. Each candidate has two paired 80-token trials. Changes compare means over two runs per variant; positive energy changes denote higher trimmed-window metrics. Both windows omit 2 s at each end but retain 80 tokens as denominator.}
\label{tab:wall_power}
\begin{adjustbox}{max width=\linewidth}
\begin{tabular}{lrrrr}
\toprule
Candidate & Token-exact trials & Decode tokens/s delta & Full J/token delta & Decode J/token delta \\
\midrule
$K=32$ & 2/2 & $-$2.891\% & $-$1.490\% & $+$3.571\% \\
$K=64$ & 2/2 & $-$3.601\% & $+$2.650\% & $+$4.905\% \\
\bottomrule
\end{tabular}
\end{adjustbox}
\end{table}

This experiment records no reduction in the trimmed decode-window metric for either candidate. Its two pairs and integer-watt readings establish the reported observations, without resolving small physical power differences. The 0.5-threshold projection trace and the 4.0-threshold full-head power runs do not isolate the same computation; the trace shares therefore cannot identify the cause of the wall-power result. Their joint practical message is that candidate-row counts and measured runtime/energy describe distinct properties of a configuration.

\FloatBarrier
\section{Conclusion}

The runtime maps binary spike activations to active-column reads and integer weight sums, while retaining row-major dot products for continuous inputs. In the early-checkpoint comparison, INT8 achieves 2.37 times the FP32 decode rate with weight storage reduced to 32.4\%; INT4mix exchanges a further 17.4\% storage reduction for 46.6\% lower decode throughput. The final 874M-parameter INT8 checkpoint reaches 22.63 tokens/s on one CPU thread and 47.90 on four with 1087.4 MiB of weights. Eight threads raise 512-token prefill to 94.68 tokens/s. These operating points characterize the CPU deployment enabled by the implemented projection layouts and quantized kernels. The separate ARM case study records no trimmed decode-window energy benefit for the tested full-head verification configurations.

\section*{Artifact Availability}

The author retains the benchmark logs, scripts, power traces, model exports, and numerical-check references supporting these measurements. CPU and prefill statistics, trace aggregates, and trimmed-window metrics can be recalculated from those records. The exact runtime and model artifacts used in the measurements are not publicly released with this paper.

\appendix
\FloatBarrier
\section{Historical Layout Measurements and Export Correction}
\label{app:history}

\subsection{FP32 Layout Measurements}

Table~\ref{tab:layout_iter} records the single-thread development comparison on the early checkpoint. SIMD denotes single-instruction, multiple-data vectorization. The scalar build uses plain C++; the following builds change vectorization and layout. Applying column-major storage to every matrix lowers throughput below the scalar baseline, while assigning it only to the sparse projections reaches 13.0 tokens/s. The development report attributes the all-column-major slowdown to repeated read--modify--write traffic on the 57,344-entry output vector, which exceeds the L1 cache; no hardware-counter measurements accompany that explanation.

\begin{table}[!htbp]
\centering
\caption{Historical FP32 layout iterations on the early checkpoint, transcribed from the development report. Single thread; decode latency and rate are per-token figures from that report.}
\label{tab:layout_iter}
\begin{adjustbox}{max width=\linewidth}
\begin{tabular}{lrrrr}
\toprule
Iteration & Prefill (2 tokens) & Decode ms/token & Decode tokens/s & vs.\ scalar \\
\midrule
Scalar FP32 (no SIMD) & 204 ms & 105 & 9.5 & 1.00$\times$ \\
AVX2, all row-major & 157 ms & 96 & 10.4 & 1.09$\times$ \\
AVX2, all column-major & 167 ms & 122 & 8.2 & 0.86$\times$ \\
AVX2, column-major + L1 tiling & 185 ms & 113 & 8.8 & 0.93$\times$ \\
AVX2, mixed layout & 135 ms & 76.8 & 13.0 & 1.37$\times$ \\
\bottomrule
\end{tabular}
\end{adjustbox}
\end{table}

Table~\ref{tab:evolution} preserves a later development session, which records 14.7 tokens/s for mixed-layout FP32. A separate sustained 100-token measurement in those reports gives 10.9 tokens/s for FP32 and 18.0 for INT8. The builds and timing sessions differ from Table~\ref{tab:layout_iter} and from the precision comparison in the main text; their rates are retained as separate measurements, not pooled into a common baseline. The historical GB unit labels are unchanged. The later precision comparison includes neither a rebuilt scalar binary nor a row-major-only variant, so it provides no repeated measurement of the original layout comparison.

\begin{table}[!htbp]
\centering
\caption{Historical runtime progression on the early checkpoint, transcribed from the INT8 and INT4 technical reports. GB is the historical unit label.}
\label{tab:evolution}
\begin{adjustbox}{max width=\linewidth}
\begin{tabular}{lrrr}
\toprule
Runtime variant & Weight size & Decode tokens/s & vs.\ scalar FP32 \\
\midrule
Scalar FP32 baseline & 3.49 GB & 9.5 & 1.00$\times$ \\
AVX2 mixed-layout FP32 & 3.28 GB & 14.7 & 1.55$\times$ \\
AVX2 INT8 & 1.06 GB & 19.9 & 2.09$\times$ \\
AVX2 INT4mix & 0.88 GB & 13.7 & 1.44$\times$ \\
\bottomrule
\end{tabular}
\end{adjustbox}
\end{table}

\subsection{Pure-INT4 Export Correction}
\label{sec:int4}

The earlier pure-INT4 timing and qualitative degradation results are withdrawn. That exporter quantized sparse weights after transposing the logical $[M,K]$ matrix to $[K,M]$, while the runtime indexed group scales as $[M,\lceil K/G\rceil]$. All 22 non-square feed-forward down matrices have scale shape $[6144,48]$ instead of $[1536,192]$ for group size $G=32$. The element counts coincide, so file-size checks do not expose the layout mismatch. INT4mix retains INT8 on these sparse projections and therefore avoids this particular defect.

\FloatBarrier
\section{Dense CPU References and Complete Decode Data}
\label{app:dense}

Dense references use llama.cpp commit \texttt{aa00911}, its CPU backend, and GGUF Q8\_0 weights (8-bit values with per-block scales) for Qwen2.5~\cite{qwen25}, TinyLlama~\cite{tinyllama}, and Falcon3~\cite{falcon3}. The harness invokes \texttt{llama-bench} with prompt length 11 and generation length 100, recording synthetic token-generation timing. It uses the same Ryzen host and three runs per setting, but differs from spiking-model text generation in prompt content, sampling work, and context history. Dense runs do not emit generated token IDs. Thread counts are passed directly to the benchmark tool.

Table~\ref{tab:cpu} retains single-thread rates, memory, and separately recorded perplexities. These references are not matched-workload speedup or quality comparisons: tokenizers differ, and dense perplexities are transcribed from their corresponding prior evaluations. Table~\ref{tab:thread_scaling} includes all decode medians and ranges, including the spiking-model data plotted in the main text.

\begin{table}[!htbp]
\centering
\caption{Single-thread CPU 8-bit inference on AMD Ryzen 7 5800X. Decode is the median over three runs with the min--max range: spiking-model text generation versus dense llama.cpp benchmark-tool probes. Perplexity (PPL) values use different tokenizers and protocols and are not directly comparable. Weight size and sampled peak RSS are MiB.}
\label{tab:cpu}
\begin{adjustbox}{max width=\linewidth}
\begin{tabular}{lrrrrrr}
\toprule
Model & Params & WikiText-2 PPL & Decode tokens/s & Range & Weight MiB & RSS MiB \\
\midrule
Spiking model, C++ INT8 & 874M & 24.80 & 22.63 & 22.42--22.83 & 1087.4 & 1175.9 \\
Qwen2.5-0.5B, GGUF Q8\_0 & 494M & 15.44 & 32.76 & 32.13--32.93 & 506.5 & 565.0 \\
TinyLlama-1.1B, GGUF Q8\_0 & 1.1B & 8.51 & 16.31 & 16.03--16.70 & 1115.6 & 1071.7 \\
Falcon3-1B, GGUF Q8\_0 & 1.0B & 11.53 & 11.26 & 11.01--11.55 & 1696.3 & 1487.1 \\
Qwen2.5-1.5B, GGUF Q8\_0 & 1.54B & 10.18 & 9.70 & 9.61--9.75 & 1570.3 & 1634.4 \\
\bottomrule
\end{tabular}
\end{adjustbox}
\end{table}

\begin{table}[!htbp]
\centering
\caption{CPU decode thread scaling on AMD Ryzen 7 5800X, tokens/s. Medians over three runs (spiking-model generation; dense benchmark-tool probes); the indented row gives the min--max of the three trials. The spiking model emitted identical generated token IDs across trials at every thread count.}
\label{tab:thread_scaling}
\begin{adjustbox}{max width=\linewidth}
\begin{tabular}{lrrrrr}
\toprule
Model & 1 thread & 2 threads & 4 threads & 8 threads & 16 threads \\
\midrule
Spiking model, C++ INT8 & 22.63 & 39.08 & 47.90 & 47.28 & 41.80 \\
\quad range & 22.42--22.83 & 38.57--40.05 & 47.19--50.00 & 46.89--47.68 & 41.76--42.35 \\
Qwen2.5-0.5B, GGUF Q8\_0 & 32.76 & 48.84 & 56.97 & 55.75 & 43.01 \\
\quad range & 32.13--32.93 & 48.29--49.66 & 56.66--57.37 & 51.61--56.84 & 40.87--43.15 \\
TinyLlama-1.1B, GGUF Q8\_0 & 16.31 & 22.74 & 25.81 & 22.99 & 17.40 \\
\quad range & 16.03--16.70 & 22.01--23.25 & 24.11--26.38 & 22.37--24.79 & 16.73--20.42 \\
Falcon3-1B, GGUF Q8\_0 & 11.26 & 15.49 & 18.53 & 18.45 & 14.90 \\
\quad range & 11.01--11.55 & 15.49--16.23 & 18.08--19.76 & 15.84--19.00 & 14.08--16.69 \\
Qwen2.5-1.5B, GGUF Q8\_0 & 9.70 & 14.32 & 16.93 & 13.72 & 10.77 \\
\quad range & 9.61--9.75 & 14.26--14.34 & 13.94--17.24 & 12.89--16.71 & 9.75--13.75 \\
\bottomrule
\end{tabular}
\end{adjustbox}
\end{table}

\FloatBarrier
\section{Complete Prefill Measurements}
\label{app:prefill}

Table~\ref{tab:long_prefill} gives the final-checkpoint synthetic-prompt measurements underlying Figure~\ref{fig:prefill}. Inputs and repetitions follow Section~\ref{sec:setup}.

\begin{table}[!htbp]
\centering
\caption{Long-prompt prefill scaling for the final INT8 export on AMD Ryzen 7 5800X, tokens/s. Medians over three runs with exact-length synthetic token IDs; the indented row gives the min--max of the three trials.}
\label{tab:long_prefill}
\begin{adjustbox}{max width=\linewidth}
\begin{tabular}{rrrrr}
\toprule
Prompt tokens & 1 thread & 2 threads & 4 threads & 8 threads \\
\midrule
16 & 31.92 & 59.20 & 95.21 & 130.89 \\
\quad range & 31.55--33.17 & 54.68--59.38 & 82.13--95.98 & 128.96--132.28 \\
64 & 33.46 & 60.68 & 107.37 & 126.35 \\
\quad range & 31.92--33.54 & 60.29--61.74 & 106.99--110.99 & 126.05--128.98 \\
128 & 32.95 & 60.31 & 95.75 & 116.16 \\
\quad range & 32.50--33.96 & 55.88--60.32 & 92.72--96.87 & 92.43--126.20 \\
256 & 33.07 & 55.68 & 96.66 & 111.57 \\
\quad range & 32.23--33.24 & 51.80--58.45 & 89.86--100.14 & 109.84--115.44 \\
512 & 29.86 & 54.87 & 77.69 & 94.68 \\
\quad range & 29.16--30.30 & 53.15--55.40 & 62.80--86.81 & 87.47--94.78 \\
\bottomrule
\end{tabular}
\end{adjustbox}
\end{table}

\FloatBarrier
\section{ARM Trace and Per-Run Power Measurements}
\label{app:power}

Table~\ref{tab:zerotrace} describes the projection-only gate at threshold 0.5. Its timing columns aggregate only the 54 events in which candidate verification ran. Table~\ref{tab:wall_abs} describes the separate full-head power configuration at threshold 4.0; the power experiment additionally configures activation outlier handling for the head projections. This is a comparison of complete configurations, not an isolated row-skipping ablation.

\begin{table}[!htbp]
\centering
\caption{ARM projection-only verification trace, gate threshold 0.5. Timing shares and total head time cover the 54 verified events out of 80. Skipped MAC (multiply-accumulate) percentage is $100(1-K/57{,}344)$ for verification rows, excluding the initial full output projection. The verification share is the remaining cost after selection.}
\label{tab:zerotrace}
\begin{adjustbox}{max width=\linewidth}
\begin{tabular}{lrrrrr}
\toprule
$K$ & Verified rows & Verification share of head time & Skipped MAC & Output projection share & Head time (ms) \\
\midrule
16 & 864 & 0.871\% & 99.972\% & 58.08\% & 1012.316 \\
32 & 1728 & 0.998\% & 99.944\% & 59.78\% & 1055.754 \\
64 & 3456 & 1.103\% & 99.888\% & 62.08\% & 1129.754 \\
\bottomrule
\end{tabular}
\end{adjustbox}
\end{table}

\begin{table}[!htbp]
\centering
\caption{Per-run ARM wall-power measurements for the full-head gate at threshold 4.0, in execution order. The runtime decode rate covers 80 tokens. $\bar P$ is the mean in-window plug reading; J/token denotes the trimmed-window metric defined in Section~\ref{sec:setup_power}. The plug reports integer watts; each decode window contains 18--20 samples.}
\label{tab:wall_abs}
\begin{adjustbox}{max width=\linewidth}
\begin{tabular}{llrrrrrr}
\toprule
$K$ & Run & Decode tokens/s & Decode $\bar P$ (W) & Decode J/token & Full $\bar P$ (W) & Full J/token & Tokens exact \\
\midrule
32 & baseline 1 & 4.727 & 8.00 & 1.2925 & 7.94 & 3.2791 & -- \\
32 & candidate 1 & 4.576 & 7.95 & 1.3399 & 7.63 & 3.2515 & yes \\
32 & candidate 2 & 4.585 & 8.00 & 1.3448 & 7.61 & 3.2055 & yes \\
32 & baseline 2 & 4.707 & 8.00 & 1.2998 & 7.88 & 3.2757 & -- \\
\midrule
64 & baseline 1 & 4.795 & 8.00 & 1.2684 & 7.51 & 3.0692 & -- \\
64 & candidate 1 & 4.606 & 8.00 & 1.3369 & 7.67 & 3.2126 & yes \\
64 & candidate 2 & 4.600 & 8.00 & 1.3390 & 7.66 & 3.2035 & yes \\
64 & baseline 2 & 4.755 & 8.00 & 1.2823 & 7.73 & 3.1812 & -- \\
\bottomrule
\end{tabular}
\end{adjustbox}
\end{table}

Individual paired changes in the decode-window metric are $+3.67\%$/$+3.47\%$ for $K=32$ and $+5.40\%$/$+4.42\%$ for $K=64$. The recorded decode means remain near 8 W, while full-window means also include loading, prefill, and process transitions. For $K=32$, full-window means are 7.63/7.61 W for candidates and 7.94/7.88 W for baselines. These full-window mixtures differ from decode-only measurements.

\FloatBarrier
\section{GPU Context Measurement}
\label{app:gpu}

Table~\ref{tab:gpu} retains the original measurement on one NVIDIA RTX 2080 Ti. It reports median decode timing over three 100-token runs and peak allocated GPU memory in MiB. Dense references use bitsandbytes LLM.int8() and were instruct/chat variants according to the benchmark notes. The spiking model uses per-channel INT8. The BitNet row is \texttt{microsoft/bitnet-b1.58-2B-4T}~\cite{bitnet}, about 2.4B parameters, using its native ternary weights. Exact model revisions are not pinned in the summaries. These GPU records are separate from the C++ CPU evaluation.

The previously reported GPU power values are omitted: they came from one \texttt{nvidia-smi} query after the timing trials, rather than a decode-window power measurement.

\begin{table}[!htbp]
\centering
\caption{GPU context measurement on RTX 2080 Ti (version 1 record). Weight formats are INT8 except for BitNet's native ternary weights. Perplexities use each model's own tokenizer and are not comparable across rows.}
\label{tab:gpu}
\begin{adjustbox}{max width=\linewidth}
\begin{tabular}{lrrrr}
\toprule
Model & Params & WikiText-2 PPL & Decode tokens/s & GPU memory MiB \\
\midrule
Spiking model, per-channel INT8 & 874M & 24.66 & 11.9 & 1271 \\
Qwen2.5-0.5B, bitsandbytes INT8 & 494M & 14.41 & 5.4 & 614 \\
TinyLlama-1.1B, bitsandbytes INT8 & 1.1B & 8.07 & 7.3 & 1199 \\
Falcon3-1B, bitsandbytes INT8 & 1.0B & 10.46 & 7.8 & 2126 \\
BitNet-b1.58-2B-4T, native ternary & 2.4B & 17.23 & 6.7 & 1520 \\
Qwen2.5-1.5B, bitsandbytes INT8 & 1.54B & 9.65 & 4.5 & 1716 \\
\bottomrule
\end{tabular}
\end{adjustbox}
\end{table}

\FloatBarrier

\end{document}